\documentclass[letterpaper,journal]{IEEEtran}

\usepackage{mathtools}
\usepackage{amssymb}
\usepackage{amsthm}

\usepackage{algorithm}
\usepackage{algorithmic}

\usepackage{array}
\usepackage{booktabs}
\usepackage{multirow}
\usepackage{graphicx}
\usepackage[
    caption=false,
    font=normalsize,
    labelfont=sf,
    textfont=sf
]{subfig}

\usepackage{textcomp}
\usepackage{stfloats}
\usepackage{url}
\usepackage{verbatim}
\usepackage{cite}
\usepackage{xcolor}

\theoremstyle{plain}

\theoremstyle{remark}

\begin{document}

\title{Constraint-Aware Physics-Informed Neural Networks for Static Shape Estimation of Co-Manipulative Continuum Robots}

\author{Rana Danesh, Pari Qarehdaghi, and Farrokh Janabi-Sharifi}

\IEEEpubid{}

\maketitle

\begin{abstract}
Static shape estimation of co-manipulative continuum robots (CCRs) is challenging because the continuum arms and manipulated flexible object form a closed chain that must satisfy both static equilibrium and geometric loop-closure constraints. This paper presents a constraint-aware physics-informed neural network (PINN) for static shape estimation of a tendon-driven CCR modeled using the geometric variable strain formulation. The proposed method incorporates a projected static equilibrium residual and a configuration-level geometric residual to enforce the governing mechanics and closed-chain geometry. In simulation, the PINN is compared with a purely data-driven artificial neural network (ANN) under limited and noisy training data. With 140 samples and 50\% label noise, the PINN reduces the relative configuration error, equilibrium residual, and closed-chain residual by 67.88\%, 67.35\%, and 88.06\%, respectively. Using the full dataset, the PINN achieves 0.1597\% relative configuration error with an inference time of 0.1773~ms, compared with 17.97~s for an iterative nonlinear solver. Experimental fine-tuning reduces the marker RMSE from 2.657~mm to 0.497~mm and increases $R^2$ from -0.788 to 0.937. These results demonstrate accurate, physically consistent, and computationally efficient static shape estimation of closed-chain CCRs.
\end{abstract}


\section{Introduction}
\label{sec:introduction}

Continuum robots (CRs) achieve complex motions through the continuous deformation of compliant backbones, enabling them to navigate confined and geometrically complex environments in which conventional rigid link manipulators may have limited accessibility \cite{BurgnerKahrs2015,WebsterJones2010,danesh2026HVS}. However, this compliance can limit stiffness, payload capacity, and positioning accuracy during manipulation. Collaborative configurations involving multiple CRs can provide additional structural support and enable manipulation tasks that may exceed the capabilities of an individual robot \cite{Lotfavar2018,Jalali2022,Li2026Survey,danesh2026active}.

Co-manipulative continuum robots (CCRs) represent one form of collaborative configuration in which multiple CRs are mechanically coupled through a shared manipulated object \cite{Chikhaoui2018,Jalali2024,Lilge2024Survey}. The manipulated object and the continuum arms together form a closed-chain mechanism. Consequently, the feasible configurations of the individual arms are not independent but must satisfy the geometric constraints imposed at their connection points. The coupling also generates internal reaction forces and moments that maintain the closed-chain constraints. These characteristics make static modeling of CCRs substantially more challenging than that of an individual open-chain CR \cite{Lilge2023Kinetostatic,Danesh2026}.

Model-based formulations can capture the distributed deformation and coupled behavior of CCRs. Among them, Cosserat rod models provide a geometrically exact representation of bending, torsion, shear, and axial deformation \cite{JanabiSharifi2021Tutorial,Armanini2023Overview,danesh2025backstepping}. The geometric variable strain (GVS) formulation provides a finite-dimensional representation of Cosserat rod mechanics by expanding the continuous strain fields over selected basis functions \cite{Renda2020GVS,Boyer2021,Mathew2023SoRoSim}. Using this formulation, the continuum arms and manipulated object can be described through a unified set of generalized coordinates. However, determining the static configuration for a prescribed tendon input requires solving a constrained nonlinear equilibrium problem. Repeating this solution for new inputs can be computationally expensive, limiting its use in applications requiring rapid model evaluations, such as real-time prediction, optimization, and control \cite{Bensch2024}.

To reduce this computational burden, data-driven surrogate models can
learn mappings from actuation variables to robot configurations using
simulation generated data or experimental measurements of the robot
state, such as position, orientation, or shape obtained from optical
markers, electromagnetic trackers, flex sensors, or fiber Bragg grating
sensors \cite{Chen2025DataDriven}. Once trained, an artificial neural network (ANN) can evaluate this mapping much faster than an iterative nonlinear solver. However, purely data-driven models do not inherently enforce the governing mechanics and may therefore produce predictions that violate static equilibrium or closed-chain constraints, particularly for unseen inputs or noisy training data. In addition, collecting a large and representative experimental dataset for CCRs can be costly and time-consuming \cite{Bensch2024,Chen2025DataDriven,Karniadakis2021}.

Physics-informed neural networks (PINNs) provide a means of combining data-driven learning with governing physical principles by incorporating equations and constraints into the learning objective \cite{Raissi2019,Karniadakis2021,Cuomo2022}. In this way, data guide the learned input--output mapping, while physics-based residuals penalize mechanically inconsistent predictions, reducing dependence on data and improving physical consistency \cite{Liu2024PINNRobotics}. Physics-informed learning has been investigated for several CR problems, including static model approximation, inverse kinematics, shape reconstruction, deformation modeling, and model predictive control \cite{Bensch2024,Lin2025,Feizi2026,Wang2024PINNRay,Licher2026,Habich2026}. However, existing studies have mainly considered individual CRs and have not addressed the simultaneous equilibrium and geometric constraints arising from the closed-chain coupling of multiple continuum arms and a shared flexible object.

In this paper, we develop a constraint-aware PINN for static shape estimation of closed-chain CCRs. The proposed formulation incorporates both static equilibrium and geometric loop-closure constraints into the learning process, enabling the network to predict physically consistent equilibrium configurations from tendon actuation inputs.

The main contributions of this work are:
\begin{itemize}

\item Unlike existing physics-informed learning approaches that primarily
consider individual continuum or soft robots
\cite{Bensch2024,Lin2025,Feizi2026,Wang2024PINNRay,Licher2026,Habich2026},
we develop a constraint-aware PINN for static shape estimation of
closed-chain CCRs. The proposed formulation simultaneously enforces
static equilibrium in the constraint-consistent subspace and geometric
loop closure through a configuration-level constraint residual.

\item A numerical assessment of the proposed PINN against a purely data-driven ANN under limited and noisy training data, as well as against an iterative nonlinear static solver, evaluating prediction accuracy, physical consistency, robustness, and computational efficiency.

\item A simulation to experiment transfer framework in which the PINN is pretrained using simulation data and subsequently fine-tuned using measured tendon displacements and marker positions from the physical CCR platform.

\end{itemize}

\section{Static Geometric Variable Strain Modeling}
\label{sec:static_gvs}
%======================================================================

\begin{figure}
    \centering
    \includegraphics[
        width=0.4\textwidth,
        trim=6cm 2.5cm 6.5cm 2cm,
        clip
    ]{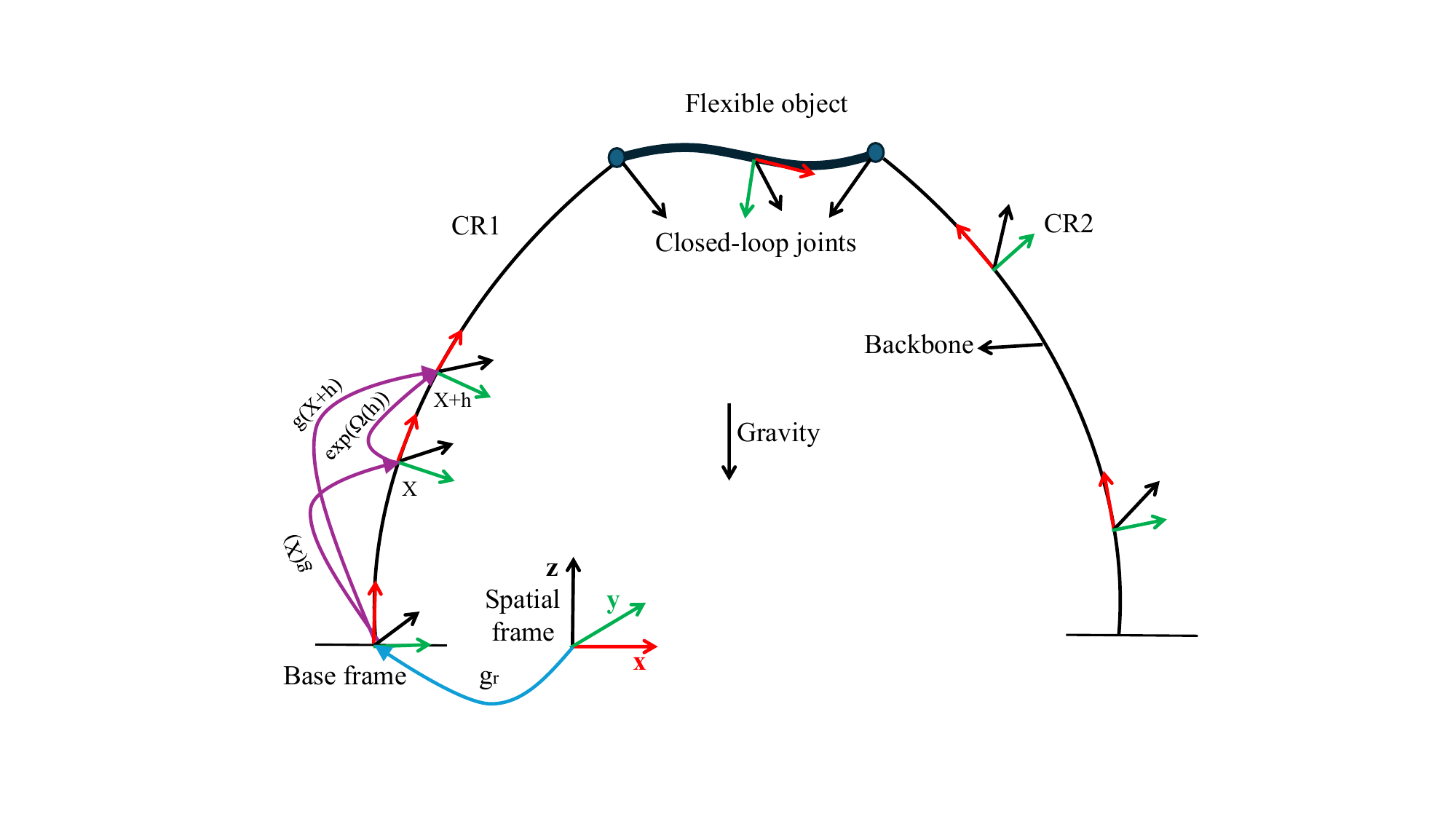}
    \caption{Tendon-driven CCR coupled through a flexible object,
    showing the backbone coordinates and reference frames \cite{Danesh2026}.}
    \label{fig:schematic}
\end{figure}

Fig.~\ref{fig:schematic} illustrates the tendon-driven
CCR considered in this study.
The system consists of two CRs mechanically coupled
through a flexible object, forming a closed kinematic chain.
The governing equations presented in this section follow the GVS formulation developed in
\cite{Renda2020GVS,renda2018discrete,Armanini2021ClosedChain,Boyer2021}
and are specialized to the static closed-chain CCR considered here.
Accordingly, only equilibrium configurations are considered, with the
generalized velocities and accelerations set to zero; thus, inertial and
damping terms are excluded from the governing model.

\subsection{GVS Kinematics}

A deformable rod is parameterized by the material coordinate
{\small $X\in[0,L]$}. As illustrated in Fig.~\ref{fig:schematic},
a local frame is attached to each cross section along the rod.
The pose of this local frame with respect to the reference frame is
represented by

{\small
\begin{equation}
\mathbf{g}(X)=
\begin{bmatrix}
\mathbf{R}(X) & \mathbf{p}(X)\\
\mathbf{0}^{\mathsf T} & 1
\end{bmatrix}
\in SE(3),
\label{eq:g_SE3_static}
\end{equation}
}

\noindent
where {\small $SE(3)$} denotes the special Euclidean group,
{\small $\mathbf{p}(X)\in\mathbb{R}^{3}$} denotes the position
of the cross-section origin, and
{\small $\mathbf{R}(X)\in SO(3)$} describes its orientation, with
{\small $SO(3)$} denoting the special orthogonal group.
The spatial evolution of the pose is governed by

{\small
\begin{equation}
\mathbf{g}'(X)
=
\mathbf{g}(X)
\widehat{\boldsymbol{\xi}}(X),
\label{eq:static_cosserat_kinematics}
\end{equation}
}

\noindent
where {\small $(\cdot)'$} denotes differentiation with respect to {\small $X$},
{\small $\widehat{(\cdot)}:\mathbb{R}^{6}\rightarrow\mathfrak{se}(3)$}
is the standard hat operator, with {\small $\mathfrak{se}(3)$} denoting
the Lie algebra of the special Euclidean group {\small $SE(3)$}, and
{\small
$\boldsymbol{\xi}(X)
=
\begin{bmatrix}
\boldsymbol{\kappa}^{\mathsf T}(X) &
\boldsymbol{\gamma}^{\mathsf T}(X)
\end{bmatrix}^{\mathsf T}
\in\mathbb{R}^{6}$
}
is the strain twist. The vectors
{\small $\boldsymbol{\kappa}\in\mathbb{R}^{3}$} and
{\small $\boldsymbol{\gamma}\in\mathbb{R}^{3}$} contain the angular and
linear strain components, respectively, accounting for bending,
torsion, shear, and axial deformation.

In the GVS formulation, the continuous strain field is
approximated using a finite set of spatial basis functions:

{\small
\begin{equation}
\boldsymbol{\xi}(X)
=
\mathbf{B}_{\boldsymbol{\xi}}(X)\mathbf{q}
+
\boldsymbol{\xi}^{\ast}(X),
\label{eq:gvs_strain_parameterization}
\end{equation}
}

\noindent
where
{\small $\mathbf{B}_{\boldsymbol{\xi}}(X)\in\mathbb{R}^{6\times n}$}
is the strain basis matrix,
{\small $\mathbf{q}\in\mathbb{R}^{n}$} is the vector of generalized
GVS coordinates,
{\small $n$} is the total number of generalized degrees of freedom,
and
{\small $\boldsymbol{\xi}^{\ast}(X)$} denotes the strain field of the
reference configuration.

The nonlinear pose field is reconstructed from
\eqref{eq:static_cosserat_kinematics}. As illustrated in
Fig.~\ref{fig:schematic}, for a spatial interval of length
{\small $h$}, the transformation is evaluated through the exponential map as

{\small
\begin{equation}
\mathbf{g}(X+h)
=
\mathbf{g}(X)
\exp\!\left(
\widehat{\boldsymbol{\Omega}}(X,h)
\right),
\label{eq:magnus_update_static}
\end{equation}
}

\noindent
where {\small $\boldsymbol{\Omega}$} denotes the Magnus increment.
Using the fourth-order Zanna collocation scheme with
two-stage Gauss quadrature gives
{\small
$\boldsymbol{\Omega}(X,h)
={}
\frac{h}{2}
\left(
\boldsymbol{\xi}_{1}
+
\boldsymbol{\xi}_{2}
\right)
+
\frac{\sqrt{3}h^{2}}{12}
\operatorname{ad}_{\boldsymbol{\xi}_{1}}
\boldsymbol{\xi}_{2},$
} where
{\small
$
\boldsymbol{\xi}_{1}
=
\boldsymbol{\xi}
\left(
X+\frac{h}{2}
-\frac{\sqrt{3}h}{6}
\right),
\label{eq:zanna_point1}
\boldsymbol{\xi}_{2}
=
\boldsymbol{\xi}
\left(
X+\frac{h}{2}
+\frac{\sqrt{3}h}{6}
\right).
$
}
where {\small $\operatorname{ad}_{\boldsymbol{\xi}}$} denotes the Lie-algebra adjoint operator associated with the strain twist {\small $\boldsymbol{\xi}$}.
This spatial integration procedure is used to reconstruct the
configuration of the continuum arms and the flexible object from
the generalized coordinates.

The geometric Jacobian is required to evaluate the generalized gravitational loads and closed-chain constraints. For the GVS parameterization, it is given by

{\small
\begin{equation}
\mathbf{J}(\mathbf{q},X)
=
\operatorname{Ad}_{\mathbf{g}^{-1}(X)}
\int_{0}^{X}
\operatorname{Ad}_{\mathbf{g}(s)}
\mathbf{B}_{\boldsymbol{\xi}}(s)\,ds,
\label{eq:gvs_geometric_jacobian_static}
\end{equation}
}

\noindent
where
{\small $\operatorname{Ad}_{\mathbf{g}}\in\mathbb{R}^{6\times6}$}
denotes the adjoint representation of {\small $SE(3)$}.

%======================================================================
\subsection{Static Equilibrium}
\label{subsec:static_equilibrium}
%======================================================================

At static equilibrium, the internal elastic wrench is related to
the strain deviation from the reference configuration through a
linear elastic constitutive law,

{\small
\begin{equation}
\boldsymbol{\mathcal F}_{e}(X)
=
\boldsymbol{\mathcal K}(X)
\left[
\boldsymbol{\xi}(X)
-
\boldsymbol{\xi}^{\ast}(X)
\right],
\label{eq:elastic_wrench_static}
\end{equation}
}

\noindent
where
{\small $\boldsymbol{\mathcal K}(X)\in\mathbb{R}^{6\times6}$}
is the cross sectional stiffness matrix. For an isotropic rod,
{\small
$
\boldsymbol{\mathcal K}
=
\operatorname{diag}
\left(
GJ_x,\,
EJ_y,\,
EJ_z,\,
EA,\,
GA,\,
GA
\right),
$
} where {\small $E$} and {\small $G$} denote Young's and shear moduli,
{\small $A$} is the cross section area, and
{\small $J_x$}, {\small $J_y$}, and {\small $J_z$} are the corresponding second
moments of area.

Substituting the strain parameterization
\eqref{eq:gvs_strain_parameterization} into
\eqref{eq:elastic_wrench_static}, the generalized stiffness matrix
of the $i$th deformable body is obtained as

{\small
\begin{equation}
\mathbf{K}_{i}
=
\int_{0}^{L_i}
\mathbf{B}_{\boldsymbol{\xi},i}^{\mathsf T}(X)
\boldsymbol{\mathcal K}_{i}(X)
\mathbf{B}_{\boldsymbol{\xi},i}(X)
\,dX
\in\mathbb{R}^{n_i\times n_i},
 i\in\{1,2,3\},
\label{eq:body_stiffness}
\end{equation}
}

\noindent
where {\small $n_i$} is the number of generalized coordinates
associated with the $i$th deformable body. The individual stiffness
matrices are assembled into the global generalized stiffness matrix,
denoted by {\small $\mathbf{K}\in\mathbb{R}^{n\times n}$}, where each
body contributes its corresponding stiffness block. The resulting
generalized elastic force is expressed as
{\small
$
\mathbf{Q}_{e}
=
\mathbf{K}\mathbf{q}
\in\mathbb{R}^{n},
$
}
where {\small $\mathbf{Q}_{e}$} denotes the generalized elastic force.

The generalized tendon actuation force is expressed as

{\small
\begin{equation}
\boldsymbol{\tau}_{a}
=
\mathbf{B}_{q}(\mathbf{q})\mathbf{u},
\label{eq:tendon_actuation}
\end{equation}
}

\noindent
where {\small $\mathbf{u}\in\mathbb{R}^{n_a}$} is the vector of tendon
actuation inputs and {\small $\mathbf{B}_{q}(\mathbf{q})$} is the
configuration-dependent generalized actuation matrix. Each column of
{\small $\mathbf{B}_{q}$} represents the generalized force associated
with a unit input from the corresponding tendon. The matrix incorporates
the tendon routing geometry and its variation with the robot
configuration and is evaluated from the current GVS configuration
\cite{Renda2020GVS}.

The gravitational loading is described by the gravity twist
{\small $\boldsymbol{\mathcal G}\in\mathbb{R}^{6}$}, expressed in the
spatial frame. The transformation
{\small $\mathbf{g}_{r}\in SE(3)$}, shown in
Fig.~\ref{fig:schematic}, relates the spatial frame to the base frame.
Using the geometric Jacobian, the distributed gravitational loading is
mapped to the generalized coordinates, resulting in the generalized
gravitational force

{\small
\begin{equation}
\mathbf{F}_{g}(\mathbf{q})
=
\int
\mathbf{J}^{\mathsf T}(\mathbf{q},X)
\boldsymbol{\mathcal M}(X)
\operatorname{Ad}^{-1}_{\mathbf{g}_{r}\mathbf{g}(\mathbf{q},X)}
\boldsymbol{\mathcal G}
\,dX, \; \mathbf{F}_{g}\in\mathbb{R}^{n},
\label{eq:generalized_gravity}
\end{equation}
}

\noindent
where {\small $\boldsymbol{\mathcal M}(X)$} is the screw inertia
density matrix. 
Assuming that no external wrench acts on the system other than gravity,
the generalized static equilibrium prior to incorporating the closed-chain
constraints is given by

{\small
\begin{equation}
\mathbf{K}\mathbf{q}
=
\mathbf{B}_{q}\mathbf{u}
+
\mathbf{F}_{g}.
\label{eq:open_static_equilibrium}
\end{equation}
}

\subsection{Closed-Chain Geometric Constraints and Jacobian}

The continuum arms and the flexible object are mechanically coupled to form a closed kinematic chain, which introduces geometric constraints on their configurations. As described in \cite{Armanini2021ClosedChain}, the $i$th closed-loop
connection relates two bodies, denoted by $A_i$ and $B_i$.
Let {\small $\mathbf{g}_{A_i}$} and
{\small $\mathbf{g}_{B_i}$} represent the configurations of the
corresponding body frames. The actual connection frames are generally
located at fixed offsets from these body frames. These offsets are
described by the constant transformations
{\small $\overline{\mathbf{g}}_{A_i}$} and
{\small $\overline{\mathbf{g}}_{B_i}$}. Therefore, the configurations
of the two connection frames are
{\small $\mathbf{g}_{A_i}^{c}
=\mathbf{g}_{A_i}\overline{\mathbf{g}}_{A_i}$} and
{\small $\mathbf{g}_{B_i}^{c}
=\mathbf{g}_{B_i}\overline{\mathbf{g}}_{B_i}$}.
The relative transformation between the two connection frames is then

{\small
\begin{equation}
\mathbf{g}_{A_iB_i}
=
\left(\mathbf{g}_{A_i}^{c}\right)^{-1}
\mathbf{g}_{B_i}^{c}.
\label{eq:relative_joint_transform}
\end{equation}
}

\noindent
This relative transformation describes the difference in position and orientation between the two sides of the closed-loop connection and is used to define the geometric closure error.

Geometric closure requires the constrained components of this relative
transformation to be zero. The matrix
{\small $\mathbf{B}_{p,i}\in\mathbb{R}^{6\times m_i}$} contains the
wrench directions constrained by the $i$th closed-loop connection,
where {\small $m_i$} is the number of constrained directions associated
with that connection. The corresponding configuration-level closure
error is defined as

{\small
\begin{equation}
\mathbf{e}_{i}(\mathbf{q})
=
\mathbf{B}_{p,i}^{\mathsf T}
\operatorname{Log}^{\vee}
\left(
\mathbf{g}_{A_iB_i}^{-1}(\mathbf{q})
\right)
\in\mathbb{R}^{m_i}.
\label{eq:constraint_error_single}
\end{equation}
}

\noindent
Here, {\small $\operatorname{Log}^{\vee}(\cdot)$} maps the relative
transformation in {\small $SE(3)$} to its six-dimensional twist
representation. For {\small $N_c$} closed-loop connections, the
individual closure errors are combined into the global geometric
constraint vector

{\small
\begin{equation}
\mathbf{e}(\mathbf{q})
=
\begin{bmatrix}
\mathbf{e}_{1}^{\mathsf T}(\mathbf{q}) &
\mathbf{e}_{2}^{\mathsf T}(\mathbf{q}) &
\cdots &
\mathbf{e}_{N_c}^{\mathsf T}(\mathbf{q})
\end{bmatrix}^{\mathsf T}
\in\mathbb{R}^{n_c},
\label{eq:global_constraint_error}
\end{equation}
}

\noindent
where {\small $n_c$} is the total number of scalar
geometric constraints.

The constraint Jacobian describes the differential relationship imposed by the closed-chain connections. For each connection, the body Jacobians
{\small $\mathbf{J}_{A_i}$} and {\small $\mathbf{J}_{B_i}$} are first
transformed to their respective joint frames as
{\small $\mathbf{J}_{A_i}^{c}
=\operatorname{Ad}_{\overline{\mathbf{g}}_{A_i}^{-1}}
\mathbf{J}_{A_i}$} and
{\small $\mathbf{J}_{B_i}^{c}
=\operatorname{Ad}_{\overline{\mathbf{g}}_{B_i}^{-1}}
\mathbf{J}_{B_i}$}. To compare the two Jacobians consistently,
{\small $\mathbf{J}_{A_i}^{c}$} is further expressed in the frame of
the opposite side through
{\small $\widetilde{\mathbf{J}}_{A_i}
=\operatorname{Ad}_{\mathbf{g}_{A_iB_i}^{-1}}
\mathbf{J}_{A_i}^{c}$}.
The constraint Jacobian associated with the $i$th connection is then

{\small
\begin{equation}
\mathbf{A}_{i}(\mathbf{q})
=
\mathbf{B}_{p,i}^{\mathsf T}
\left(
\widetilde{\mathbf{J}}_{A_i}
-
\mathbf{J}_{B_i}^{c}
\right)
\in\mathbb{R}^{m_i\times n}.
\label{eq:A_i_constraint}
\end{equation}
}

\noindent
The individual blocks are stacked to form the global constraint
Jacobian
{\small $\mathbf{A}(\mathbf{q})\in\mathbb{R}^{n_c\times n}$}.
Its null space defines the tangent directions that are compatible with
the closed-chain constraints.

The geometric constraint vector
{\small $\mathbf{e}(\mathbf{q})$} and the constraint Jacobian
{\small $\mathbf{A}(\mathbf{q})$} have distinct roles. The vector
{\small $\mathbf{e}(\mathbf{q})$} represents the configuration-level
closed-chain error, whereas {\small $\mathbf{A}(\mathbf{q})$} defines
the differential constraint directions and is used to account for the
unknown closed-chain reaction forces in the static equilibrium
equations.

\subsection{Constraint-Consistent Static Equilibrium}

The closed-chain connections generate internal reaction forces and
moments that maintain the geometric constraints. These reactions are
represented by the Lagrange multiplier vector
{\small $\boldsymbol{\lambda}\in\mathbb{R}^{n_c}$} \cite{Armanini2021ClosedChain}. The constrained
static equilibrium is expressed as

{\small
\begin{equation}
\mathbf{B}_{q}(\mathbf{q})\mathbf{u}
+
\mathbf{F}_{g}(\mathbf{q})
-
\mathbf{K}\mathbf{q}
+
\mathbf{A}^{\mathsf T}(\mathbf{q})\boldsymbol{\lambda}
=
\mathbf{0}.
\label{eq:static_with_lambda}
\end{equation}
}

\noindent
The unknown reaction forces need not be determined explicitly. Instead,
the equilibrium equation is projected onto the null space of the
constraint Jacobian \cite{aghili2011projection}. Assuming that
{\small $\mathbf{A}(\mathbf{q})$} has full row rank, the corresponding
orthogonal projector is

{\small
\begin{equation}
\mathbf{P}(\mathbf{q})
=
\mathbf{I}
-
\mathbf{A}^{\mathsf T}(\mathbf{q})
\left[
\mathbf{A}(\mathbf{q})
\mathbf{A}^{\mathsf T}(\mathbf{q})
\right]^{-1}
\mathbf{A}(\mathbf{q}).
\label{eq:static_projector}
\end{equation}
}

\noindent
The projector satisfies
{\small $\mathbf{P}(\mathbf{q})\mathbf{A}^{\mathsf T}(\mathbf{q})
=\mathbf{0}$}, so the unknown constraint reactions do not contribute
to the projected equilibrium. The corresponding physics residual is
defined as

{\small
\begin{equation}
\mathbf{E}_{\mathrm{phys}}(\mathbf{q},\mathbf{u})
=
\mathbf{P}(\mathbf{q})
\left[
\mathbf{B}_{q}(\mathbf{q})\mathbf{u}
+
\mathbf{F}_{g}(\mathbf{q})
-
\mathbf{K}\mathbf{q}
\right].
\label{eq:physics_residual}
\end{equation}
}

\noindent
This residual enforces static equilibrium in the
constraint-consistent directions but does not independently enforce
geometric loop closure. Therefore, it is combined with the
configuration constraint vector
{\small $\mathbf{e}(\mathbf{q})$} to form the complete static residual

{\small
\begin{equation}
\mathbf{E}(\mathbf{q},\mathbf{u})
=
\begin{bmatrix}
\mathbf{E}_{\mathrm{phys}}(\mathbf{q},\mathbf{u})\\[1mm]
\mathbf{e}(\mathbf{q})
\end{bmatrix}.
\label{eq:complete_static_residual}
\end{equation}
}

\noindent
For a prescribed tendon input {\small $\mathbf{u}$}, the corresponding
static configuration {\small $\mathbf{q}^{\ast}$} satisfies

{\small
\begin{equation}
\mathbf{E}(\mathbf{q}^{\ast},\mathbf{u})=\mathbf{0}.
\label{eq:static_solution_definition}
\end{equation}
}

\section{Physics-Informed Static Shape Estimation}
Solving \eqref{eq:static_solution_definition} with an iterative nonlinear
solver for each tendon tension input can be computationally demanding.
The proposed PINN therefore learns the mapping from the tendon tension
vector {\small $\mathbf{u}$} to the corresponding equilibrium GVS
coordinates {\small $\mathbf{q}$}.

The tendon tensions are normalized to the range {\small $[-1,1]$}
using the minimum and maximum values of the training data. For the
$k$th tendon tension,

{\small
\begin{equation}
\widetilde{u}_{k}
=
2\frac{u_{k}-u_{k,\min}}
{u_{k,\max}-u_{k,\min}}
-1,
\label{eq:input_normalization}
\end{equation}
}

\noindent
where {\small $u_{k,\min}$} and {\small $u_{k,\max}$} are the
corresponding minimum and maximum training values. The normalized
tendon tension vector is denoted by
{\small $\widetilde{\mathbf{u}}\in\mathbb{R}^{4}$}.

The generalized coordinates are standardized using the mean and standard
deviation of the training data as

{\small
\begin{equation}
\widetilde{q}_{k}
=
\frac{q_{k}-\mu_{q,k}}
{\sigma_{q,k}},
\label{eq:output_standardization}
\end{equation}
}

\noindent
where {\small $\mu_{q,k}$} and {\small $\sigma_{q,k}$} are the
corresponding training set mean and standard deviation. The PINN mapping
is then expressed as

{\small
\begin{equation}
\widehat{\widetilde{\mathbf{q}}}
=
\mathcal{N}_{\boldsymbol{\theta}}
\left(
\widetilde{\mathbf{u}}
\right),
\label{eq:pinn_mapping}
\end{equation}
}

\noindent
where {\small $\mathcal{N}_{\boldsymbol{\theta}}$} denotes the neural
network with trainable parameters {\small $\boldsymbol{\theta}$}.
The predicted standardized coordinates are transformed back to the
physical GVS coordinates {\small $\widehat{\mathbf{q}}$}, from which
the strain field and the corresponding CCR backbone configurations are
reconstructed using the GVS kinematic model.

The PINN is implemented as a fully connected feedforward neural network
with four hidden layers of {\small $128$} neurons each and hyperbolic
tangent activation functions. A linear output layer maps the normalized
tendon inputs to the {\small $16$} standardized GVS coordinates.

Since the governing formulation is static, no temporal or spatial
derivatives of the network output are required. Instead, the predicted
coordinates {\small $\widehat{\mathbf{q}}$} are passed through the
differentiable GVS model to evaluate the projected equilibrium and
closed-chain residuals. The gradient used for training is

{\small
\begin{equation}
\nabla_{\boldsymbol{\theta}}\mathcal{L}
=
\lambda_{\mathrm{data}}\nabla_{\boldsymbol{\theta}}\mathcal{L}_{\mathrm{data}}
+
\lambda_{\mathrm{phys}}\nabla_{\boldsymbol{\theta}}\mathcal{L}_{\mathrm{phys}}
+
\lambda_{\mathrm{con}}\nabla_{\boldsymbol{\theta}}\mathcal{L}_{\mathrm{con}}.
\label{eq:pinn_gradient}
\end{equation}
}

\noindent
These gradients are computed using automatic differentiation and
backpropagated through both the neural network and the differentiable
GVS operations, including the forward kinematics, tendon actuation,
constraint Jacobian, projected equilibrium residual, and geometric
constraint residual. The resulting parameter gradients are used by the
Adam optimizer to update {\small $\boldsymbol{\theta}$}.

The PINN is trained using
{\small $N_d$} labeled samples
{\small $(\widetilde{\mathbf{u}}^{(j)},
\widetilde{\mathbf{q}}^{(j)})$}.
For each normalized tendon tension input
{\small $\widetilde{\mathbf{u}}^{(j)}$}, the network predicts
{\small $\widehat{\widetilde{\mathbf{q}}}^{(j)}
=\mathcal{N}_{\boldsymbol{\theta}}
(\widetilde{\mathbf{u}}^{(j)})$}.
The supervised loss measures the difference between the predicted
and reference generalized coordinates in the standardized space as

{\small
\begin{equation}
\mathcal{L}_{\mathrm{data}}
=
\frac{1}{N_d n}
\sum_{j=1}^{N_d}
\left\|
\widehat{\widetilde{\mathbf{q}}}^{(j)}
-
\widetilde{\mathbf{q}}^{(j)}
\right\|_{2}^{2}.
\label{eq:data_loss}
\end{equation}
}

\noindent
To evaluate the physical constraints, the predicted generalized
coordinates and tendon tensions are transformed back to their physical
values and evaluated using the constrained static model developed in the
previous section. Using the {\small $N_d$} training samples, the
static equilibrium loss is defined as

{\small
\begin{equation}
\mathcal{L}_{\mathrm{phys}}
=
\frac{1}{N_d n}
\sum_{j=1}^{N_d}
\left\|
\mathbf{E}_{\mathrm{phys}}
\left(
\widehat{\mathbf{q}}^{(j)},
\mathbf{u}^{(j)}
\right)
\right\|_{2}^{2},
\label{eq:physics_loss}
\end{equation}
}

\noindent
where {\small $\mathbf{E}_{\mathrm{phys}}$} is the projected static
equilibrium residual defined in \eqref{eq:physics_residual}.

Geometric consistency of the closed chain is imposed through the
configuration-level constraint residual
{\small $\mathbf{e}(\mathbf{q})$}. The corresponding constraint loss is

{\small
\begin{equation}
\mathcal{L}_{\mathrm{con}}
=
\frac{1}{N_d n_c}
\sum_{j=1}^{N_d}
\left\|
\mathbf{e}
\left(
\widehat{\mathbf{q}}^{(j)}
\right)
\right\|_{2}^{2},
\label{eq:constraint_loss}
\end{equation}
}

\noindent
where {\small $n_c$} is the number of geometric constraint components.

The complete training objective combines the supervised data,
static equilibrium, and closed-chain constraint losses as

{\small
\begin{equation}
\mathcal{L}
=
\lambda_{\mathrm{data}}\mathcal{L}_{\mathrm{data}}
+
\lambda_{\mathrm{phys}}\mathcal{L}_{\mathrm{phys}}
+
\lambda_{\mathrm{con}}\mathcal{L}_{\mathrm{con}},
\label{eq:total_pinn_loss}
\end{equation}
}

\noindent
where {\small $\lambda_{\mathrm{data}}$},
{\small $\lambda_{\mathrm{phys}}$}, and
{\small $\lambda_{\mathrm{con}}$} scale the contributions of the
individual loss terms. The supervised loss is evaluated using
standardized GVS coordinates and is therefore dimensionless, whereas
the equilibrium and geometric constraint losses are evaluated in their
physical spaces and consequently have different numerical scales and
physical units. To prevent any single term from dominating the
optimization, the loss weights are selected based on the relative
gradient magnitudes of the individual terms, thereby balancing their
contributions to the network parameter updates.

%======================================================================
\section{Simulation Results}
\label{sec:simulation}
%======================================================================

The simulation study evaluates the proposed PINN in two stages. First,
its prediction accuracy and physical consistency are compared with those
of a data driven ANN under different training set sizes and levels of
training label noise. Second, a PINN trained using the full simulation
dataset is compared with an iterative nonlinear static solver in terms
of prediction accuracy and computation time.

%======================================================================
\subsection{Dataset Generation and Training Setup}
\label{subsec:simulation_setup}
%======================================================================

The simulation dataset was generated using the constrained static GVS
model implemented in MATLAB. For the GVS discretization, each CR was
modeled as a single soft link with one division, with linear bending
basis functions in two orthogonal directions, resulting in
{\small $4$} generalized coordinates per CR. The flexible object was
represented by {\small $2$} deformation coordinates associated with
bending in two orthogonal directions, together with {\small $6$}
spatial motion coordinates. Consequently, the complete CCR model has
{\small $n=4+4+2+6=16$} generalized coordinates,
{\small $\mathbf{q}\in\mathbb{R}^{16}$}.

The four tendon actuation inputs were independently varied from
{\small $-5$} to {\small $5~\mathrm{N}$} in increments of
{\small $1~\mathrm{N}$}. Each actuation input represents the
differential action of an antagonistic tendon pair. In the experimental
setup, the corresponding motor can rotate in either the clockwise (CW)
or counterclockwise (CCW) direction, such that one tendon of the pair
is tensioned depending on the direction of rotation. Accordingly, the
sign of the input specifies which tendon is actuated and does not imply
a physically negative or compressive tendon tension. This sampling
resulted in {\small $11^4=14\,641$} input combinations. For each tendon
actuation vector {\small $\mathbf{u}\in\mathbb{R}^{4}$}, the
corresponding equilibrium GVS coordinates were obtained by solving
\eqref{eq:static_solution_definition} using MATLAB \texttt{fsolve}
with the Levenberg--Marquardt algorithm. Multiple initial guesses were
used to improve convergence, and a solution was retained only when the
solver converged and both the static equilibrium and geometric
constraint residual norms were below {\small $10^{-5}$}.

To investigate data efficiency, training sets containing
{\small $28$}, {\small $70$}, and {\small $140$} configurations were
considered. For each case, the same training samples were used for the
ANN and PINN, while the remaining configurations were kept noise free
for evaluation.

Robustness to inaccurate training labels was evaluated using
{\small $0\%$}, {\small $25\%$}, and {\small $50\%$} Gaussian noise.
Noise was applied only to the GVS coordinates used as training labels;
the tendon tension inputs and evaluation data remained unchanged. For
the $k$th generalized coordinate, the noisy training label was generated
as

\begin{equation}
q_{k,\eta}^{(j)}
=
q_k^{(j)}
+
\varepsilon_k^{(j)},
\qquad
\varepsilon_k^{(j)}
\sim
\mathcal{N}
\left(
0,\eta^2\sigma_{q_k}^{2}
\right),
\label{eq:training_noise}
\end{equation}

\noindent
where {\small $\sigma_{q_k}$} is the standard deviation of the $k$th
generalized coordinate in the corresponding clean training set and
{\small $\eta\in\{0,0.25,0.50\}$} denotes the imposed noise level.

For a fair comparison, the ANN and PINN use the same network
architecture, initialization, and optimization settings. Each network
contains four hidden layers with {\small $128$} neurons per layer and
hyperbolic tangent activation functions, followed by a linear output
layer. The network maps four tendon tension inputs to sixteen GVS
coordinates and contains {\small $52\,240$} trainable parameters.
Both models are trained for {\small $500$} epochs using the Adam
optimizer with a learning rate of {\small $10^{-3}$}. The ANN is
trained only with the supervised data loss, whereas the PINN additionally
incorporates the projected static equilibrium and closed chain
geometric constraint losses.

The PINN loss weights were selected by balancing the gradient magnitudes of
the individual loss terms at network initialization. Specifically, let

{\small
\begin{equation}
g_i
=
\left\|
\nabla_{\boldsymbol{\theta}}\mathcal{L}_i
\right\|_2,
\qquad
i\in\{\mathrm{data},\mathrm{phys},\mathrm{con}\},
\label{eq:loss_gradient_norm}
\end{equation}
}

\noindent
where the gradient is evaluated with respect to all trainable network
parameters using the labeled training set. Taking the supervised term
as the reference, {\small $\lambda_{\mathrm{data}}=1$}, the remaining
weights are determined as

{\small
\begin{equation}
\lambda_{\mathrm{phys}}
=
\frac{g_{\mathrm{data}}}{g_{\mathrm{phys}}},
\qquad
\lambda_{\mathrm{con}}
=
\frac{g_{\mathrm{data}}}{g_{\mathrm{con}}}.
\label{eq:loss_weight_selection}
\end{equation}
}

\noindent
At initialization, the corresponding gradient norms were
{\small $g_{\mathrm{data}}=3.5530\times10^{-1}$},
{\small $g_{\mathrm{phys}}=1.2252\times10^{-4}$}, and
{\small $g_{\mathrm{con}}=1.4852\times10^{-4}$}, yielding
{\small $\lambda_{\mathrm{phys}}\approx2899.8$} and
{\small $\lambda_{\mathrm{con}}\approx2392.3$}. These values were
rounded to

{\small
\begin{equation}
\lambda_{\mathrm{data}}=1,
\qquad
\lambda_{\mathrm{phys}}=3000,
\qquad
\lambda_{\mathrm{con}}=2500.
\label{eq:simulation_loss_weights}
\end{equation}
}

%======================================================================
\subsection{PINN and ANN Comparison}
\label{subsec:noise_comparison}
%======================================================================

Table~\ref{tab:simulation_pinn_gain} summarizes the improvement of the
PINN relative to the ANN for training sets of {\small $28$},
{\small $70$}, and {\small $140$} samples under different levels of
training label noise. The reported values denote the percentage
reductions in relative configuration error, projected equilibrium
residual, and closed chain residual.

\begin{table}[!t]
\caption{PINN Improvement Relative to ANN}
\label{tab:simulation_pinn_gain}
\centering
\footnotesize
\setlength{\tabcolsep}{4pt}
\renewcommand{\arraystretch}{1.15}

\resizebox{\columnwidth}{!}{%
\begin{tabular}{ccrrr}
\toprule
\textbf{$N_d$} &
\textbf{Noise} &
\textbf{Rel. $\ell_2$ Red. (\%)} &
\textbf{$\overline{\|\mathbf{E}_{\mathrm{phys}}\|_2}$ Red. (\%)} &
\textbf{$\overline{\|\mathbf{e}\|_2}$ Red. (\%)} \\
\midrule

\multirow{3}{*}{28}
& 0\%  & 4.94  & 3.14  & 45.73 \\
& 25\% & 41.59 & 36.21 & 76.16 \\
& 50\% & 45.26 & 38.34 & 78.31 \\

\midrule

\multirow{3}{*}{70}
& 0\%  & 8.66  & 7.01  & 11.40 \\
& 25\% & 61.58 & 59.85 & 87.10 \\
& 50\% & 62.04 & 59.23 & 84.23 \\

\midrule

\multirow{3}{*}{140}
& 0\%  & 9.14  & 12.71 & 49.59 \\
& 25\% & 65.81 & 65.01 & 87.26 \\
& 50\% & 67.88 & 67.35 & 88.06 \\

\bottomrule
\end{tabular}%
}
\end{table}

\begin{figure}[!t]
    \centering
    \includegraphics[
        width=0.9\columnwidth
    ]{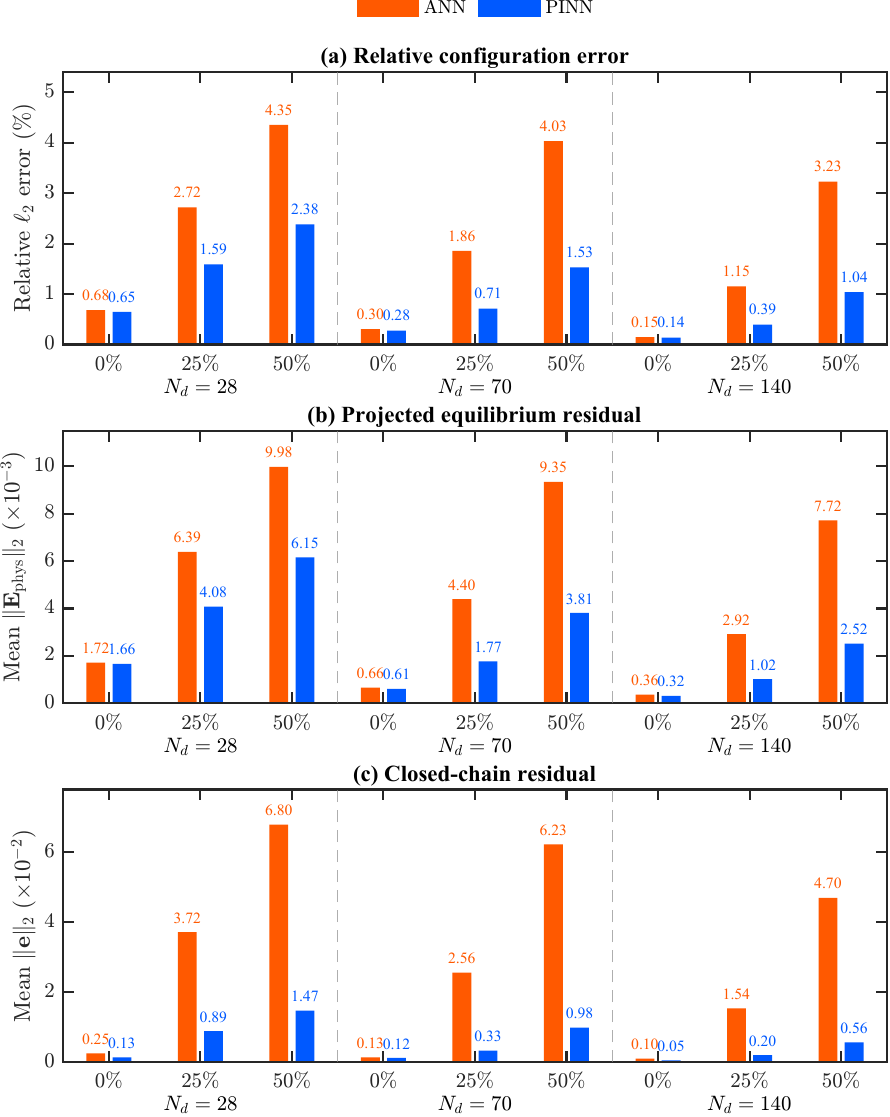}
    \caption{ANN and PINN performance for different training set sizes
    and training label noise levels: (a) relative configuration error,
    (b) mean projected equilibrium residual, and (c) mean closed chain
    residual.}
    \label{fig:ann_pinn_noise}
\end{figure}

For noise free labels, the ANN and PINN provide similar prediction
accuracy, with only modest reductions in relative error obtained by the
PINN. As the training label noise increases, however, the advantage of
the PINN becomes substantially more pronounced. At {\small $50\%$}
noise, the relative error reduction increases from
{\small $45.26\%$} with {\small $28$} training samples to
{\small $67.88\%$} with {\small $140$} samples.

Figure~\ref{fig:ann_pinn_noise} shows the corresponding absolute
performance. Increasing the training set size generally reduces the
prediction error of both models, while the PINN remains considerably
more robust to corrupted labels. For example, at
{\small $50\%$} noise, the PINN relative error decreases from
{\small $2.38\%$} with {\small $28$} samples to
{\small $1.04\%$} with {\small $140$} samples, whereas the ANN
maintains substantially larger errors.

The physical consistency measures show the same trend. Under noisy
training conditions, the PINN consistently produces lower projected
equilibrium and closed chain residuals than the ANN. With
{\small $140$} samples and {\small $50\%$} noise, these residuals are
reduced by {\small $67.35\%$} and {\small $88.06\%$}, respectively.
These results indicate that the physics based loss terms act as
regularizers, improving robustness to noisy supervision while promoting
mechanically consistent predictions.

\subsection{PINN and Nonlinear Static Solver Comparison}
\label{subsec:solver_comparison}

The computational performance of the full data PINN was compared with
an iterative nonlinear static solver using {\small $30$} configurations
selected from the held out test set, including interior, high input norm,
and boundary cases. Both methods were evaluated using the same tendon
inputs and reference GVS configurations.

The comparison was performed in Python on the CPU. The nonlinear
equilibrium equations were solved using the Levenberg--Marquardt
algorithm with a three point finite difference Jacobian, stopping
tolerances of {\small $10^{-12}$}, and a maximum of
{\small $20\,000$} function evaluations. A solution was accepted only
when the solver converged and both the complete static residual and
closed chain residual were below {\small $10^{-5}$}. To improve
convergence, the previous accepted equilibrium configuration was used
as the primary initial guess, with additional initial guesses considered
when necessary. For the PINN, each input was evaluated after
{\small $100$} warm up runs and timed over {\small $500$} repeated
forward evaluations. 

\iffalse
\begin{table}[!t]
\caption{Prediction Accuracy, Physical Consistency, and Computation Time of the PINN and Nonlinear Static Solver}
\label{tab:pinn_solver_comparison}
\centering
\footnotesize
\setlength{\tabcolsep}{3.5pt}
\renewcommand{\arraystretch}{1.15}

\resizebox{\columnwidth}{!}{%
\begin{tabular}{lrrrr}
\toprule
\textbf{Method} &
\textbf{Rel. $\ell_2$ Error (\%)} &
\textbf{$\overline{\|\mathbf{E}_{\mathrm{phys}}\|_2}$} &
\textbf{$\overline{\|\mathbf{e}\|_2}$} &
\textbf{Median Time (ms)} \\
\midrule

Direct PINN
& 0.1597
& $4.456\times10^{-4}$
& $1.437\times10^{-3}$
& \textbf{0.204} \\

Nonlinear solver
& \textbf{0.0096}
& $\mathbf{1.667\times10^{-6}}$
& $\mathbf{5.035\times10^{-6}}$
& 14259.1 \\

\bottomrule
\end{tabular}%
}
\end{table}
\fi

\begin{table}[!t]
\caption{PINN and Nonlinear Solver Benchmark}
\label{tab:pinn_solver_comparison}
\centering
\footnotesize
\setlength{\tabcolsep}{3.5pt}
\renewcommand{\arraystretch}{1.15}

\resizebox{\columnwidth}{!}{%
\begin{tabular}{lrrrr}
\toprule
\textbf{Method} &
\textbf{Rel. $\ell_2$ Error (\%)} &
\textbf{$\overline{\|\mathbf{E}_{\mathrm{phys}}\|_2}$} &
\textbf{$\overline{\|\mathbf{e}\|_2}$} &
\textbf{Median Time (ms)} \\
\midrule

Direct PINN
& 0.1597
& $4.456\times10^{-4}$
& $1.437\times10^{-3}$
& \textbf{0.1773} \\

Nonlinear solver
& \textbf{0.0053}
& $\mathbf{5.326\times10^{-8}}$
& $\mathbf{4.325\times10^{-7}}$
& 17974.2 \\

\bottomrule
\end{tabular}%
}
\end{table}

As shown in Table~\ref{tab:pinn_solver_comparison}, the nonlinear solver
provides the more accurate equilibrium solution, with a relative
{\small $\ell_2$} error of {\small $0.0053\%$} compared with
{\small $0.1597\%$} for the PINN, and produces substantially smaller
physical residuals. In contrast, the PINN requires only
{\small $0.1773~\mathrm{ms}$} per prediction, compared with
{\small $17.97~\mathrm{s}$} for the nonlinear solver. For the
implementation considered here, this corresponds to approximately
{\small $1.01\times10^{5}$} times lower median computation time while
maintaining a relative configuration error below {\small $0.16\%$}.

\section{Experimental Results}
\label{sec:exp_results}

The experimental study evaluates the proposed framework on the physical CCR system. The trained simulation PINN is first tested directly on experimental data to quantify the simulation-to-real discrepancy. It is then fine-tuned using experimental data while retaining physics based constraints and evaluated on an independent test set.

\subsection{Experimental Setup}
\label{subsec:experimental_setup}

\begin{figure}
\centering
\includegraphics[
    width=0.8\columnwidth,
    trim=0cm 0cm 0cm 0cm,
    clip]{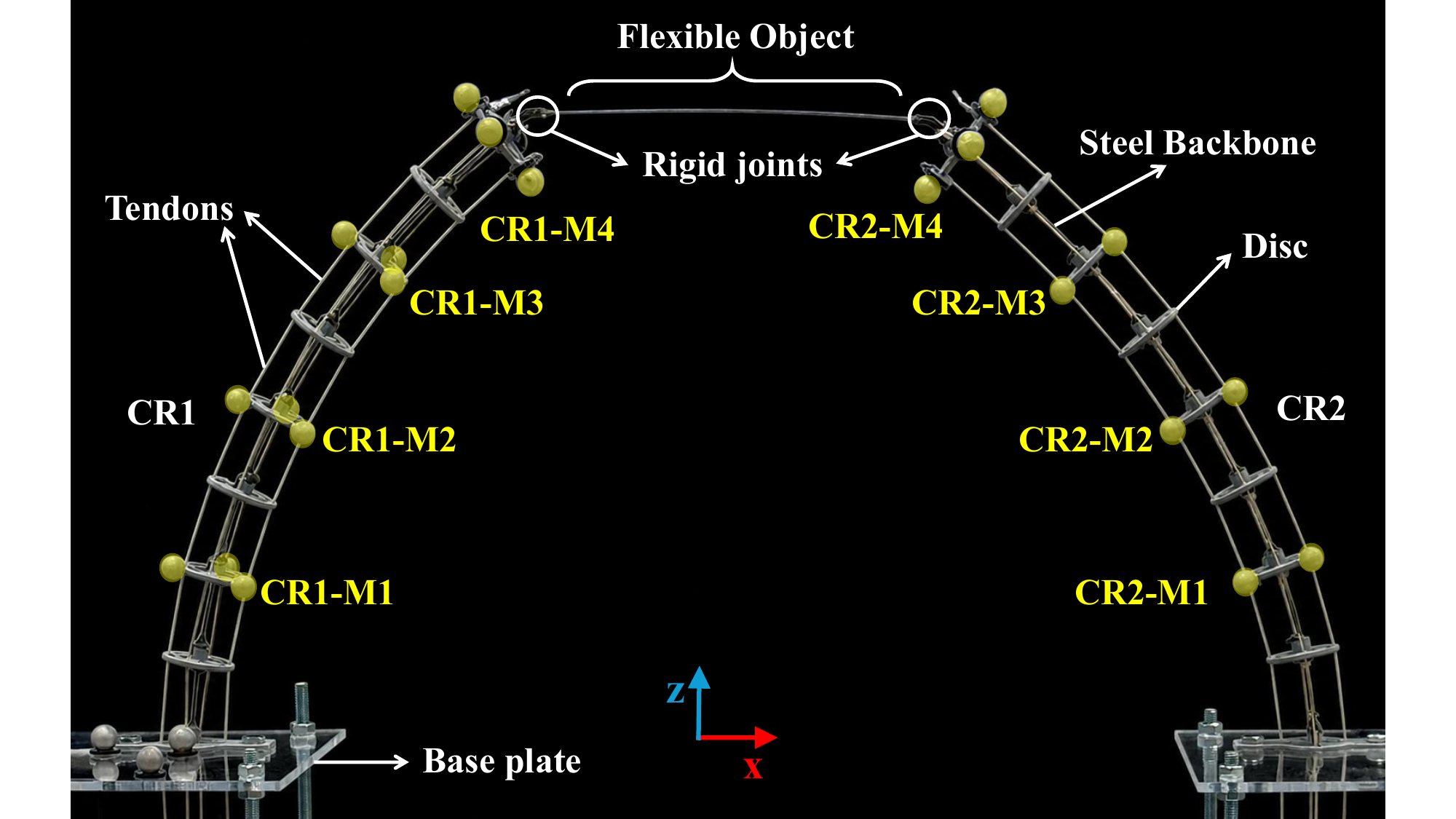}
\caption{Experimental platform of the two-arm tendon-driven CCR,
showing the continuum arms, flexible manipulated object, tendon
actuation system, and reflective markers used for configuration
measurement.}
\label{fig:exp_setup}
\end{figure}

Figure~\ref{fig:exp_setup} shows the mechanical structure of the
tendon-driven CCR platform. Each CR consists of a
{\small $0.4~\mathrm{m}$} long spring steel backbone with a diameter of
{\small $1.8~\mathrm{mm}$}, Young's modulus
{\small $E=207~\mathrm{GPa}$}, and Poisson's ratio
{\small $\nu=0.3$}. The material density of the steel backbone is
{\small $\rho=7800~\mathrm{kg/m^3}$}. To account for the combined mass
of the markers, glue, and spacer disks, an effective density of
{\small $\rho_{\mathrm{eff}}=39317~\mathrm{kg/m^3}$} is used in the
model, computed as
{\small $\rho_{\mathrm{eff}}=m_{\mathrm{total}}/(AL)$}.
Four Kevlar tendons pass through equally spaced spacer disks at a radial
distance of {\small $r_t=1.8~\mathrm{mm}$}. The tendons are separated
by {\small $90^\circ$} and arranged as two orthogonal pairs, allowing
spatial bending of each arm. The rigid base plates supporting the two
arms are separated by {\small $629~\mathrm{mm}$}. The manipulated
flexible object is a {\small $0.22~\mathrm{m}$} long Nitinol rod with
a diameter of {\small $1.6~\mathrm{mm}$}, density
{\small $\rho=6450~\mathrm{kg/m^3}$}, Young's modulus
{\small $E=50~\mathrm{GPa}$}, and Poisson's ratio
{\small $\nu=0.3$}. It is rigidly connected to the distal ends of the
two CRs, thereby completing the closed-chain structure.

The tendons are driven by Dynamixel AX-12A servo motors (Robotis, Seoul, Korea) through serial communication. The three-dimensional configuration of the CCR is measured using a Tracker 3.0 Vicon motion-capture system (Bilston, UK). As shown in Fig. \ref{fig:exp_setup}, three reflective markers are attached to the base plate and used as reference markers to define the global coordinate frame. In addition, the positions of eight markers distributed along the CCR are recorded, with four markers placed on CR1, denoted as CR1\_M1 to CR1\_M4, and four markers placed on CR2, denoted as CR2\_M1 to CR2\_M4. Each marker position is represented by its Cartesian coordinates, providing a spatial representation of the deformed CCR configuration.

\subsection{Static Experimental Data Collection}

The experimental dataset was collected under static conditions to remain
consistent with the equilibrium formulation in Section~\ref{sec:static_gvs}.
The commanded tendon actuation inputs were varied within
{\small $[-5,5]~\mathrm{N}$}. For each actuation input, the constrained
static GVS model was first solved to obtain the corresponding equilibrium
configuration. The tendon lengths associated with this configuration were
then computed from the tendon routing geometry along the deformed backbone
using the same GVS discretization and numerical quadrature employed in the
model. The tendon displacements were defined relative to the tendon lengths
at the zero-input equilibrium configuration. These displacements were then
converted to the corresponding motor angular displacements using the known
effective winding radius of the actuator shaft, and finally mapped to servo
encoder increments and transmitted to the motors.

After the commanded positions were reached, the CCR was allowed to settle before measurement. Since the generalized GVS coordinates {\small $\mathbf{q}$} cannot be measured directly on the physical platform, the three-dimensional positions of the eight Vicon markers were used to represent the experimental CCR configuration. The marker positions and tendon displacements were recorded for each test, resulting in {\small $817$} valid static configurations.

For data processing, the measured marker positions were expressed relative to the zero-input configuration. Each experimental sample is therefore represented as

{\small
\begin{equation}
\left(
\mathbf{u}^{(j)},
\Delta\boldsymbol{\ell}^{(j)},
\Delta\mathbf{p}_{\mathrm{exp}}^{(j)}
\right),
\label{eq:experimental_sample}
\end{equation}
}

\noindent
where {\small $\mathbf{u}^{(j)}$} denotes the commanded actuation input,
{\small $\Delta\boldsymbol{\ell}^{(j)}$} is the corresponding tendon
displacement relative to the zero-input configuration, and
{\small $\Delta\mathbf{p}_{\mathrm{exp}}^{(j)}$} contains the measured
three-dimensional displacements of the eight Vicon markers relative to
their zero-input positions.

%======================================================================
\subsection{Simulation Pretraining for Experimental Transfer}
\label{subsec:experimental_pretraining}
%======================================================================

The PINN is formulated using quantities available from the physical actuation and measurement systems. In contrast to the simulation study in Section~\ref{sec:simulation}, which considers the mapping from the actuation inputs, {\small $\mathbf{u}$}, to the generalized coordinates, {\small $\mathbf{q}$}, the experimental implementation uses the tendon displacements, {\small $\Delta\boldsymbol{\ell}$}, as the network input. The network contains a shared nonlinear feature representation followed by two output branches. The primary branch predicts the displacements of the eight Vicon markers, {\small $\Delta\widehat{\mathbf{p}}\in\mathbb{R}^{24}$}, while the second branch predicts the GVS generalized coordinates, {\small $\widehat{\mathbf{q}}\in\mathbb{R}^{16}$}. The generalized coordinates serve as an auxiliary physical state through which the GVS forward kinematics, tendon length relation, projected static equilibrium, and closed chain geometric constraints are incorporated into the learning process.

The network was pretrained using {\small $1331$} configurations generated by the static MATLAB model, including {\small $931$} training, {\small $199$} validation, and {\small $201$} test configurations. It consists of four shared hidden layers with {\small $128$} neurons per layer and hyperbolic tangent activation functions. The shared representation is followed by marker position and generalized coordinate output branches of dimensions {\small $24$} and {\small $16$}, respectively, resulting in {\small $55\,208$} trainable parameters.

Simulation pretraining consisted of an initial supervised stage followed by physics informed optimization. In the supervised stage, the network was trained using the marker position and generalized coordinate losses,

{\small
\begin{equation}
\mathcal{L}_{\mathrm{sup}}
=
w_p\mathcal{L}_{p}
+
w_q\mathcal{L}_{q},
\label{eq:simulation_supervised_loss}
\end{equation}
}

\noindent
where {\small $\mathcal{L}_{p}$} and {\small $\mathcal{L}_{q}$} denote the marker position and generalized coordinate losses, respectively. The corresponding weights were set to {\small $w_p=1$} and {\small $w_q=0.2$}.

Following the supervised stage, the loss weights were calibrated once based on the relative gradient magnitudes of the individual loss terms and then kept fixed during physics informed training. The complete objective was

{\small
\begin{equation}
\mathcal{L}_{\mathrm{pre}}
={}
w_p\mathcal{L}_{p}
+
w_q\mathcal{L}_{q}
+
w_{\mathrm{fk}}\mathcal{L}_{\mathrm{fk}}
+
w_{\ell}\mathcal{L}_{\ell}
+
w_{\mathrm{eq}}\mathcal{L}_{\mathrm{phys}}
+
w_{\mathrm{con}}\mathcal{L}_{\mathrm{con}},
\label{eq:simulation_pretraining_loss}
\end{equation}
}

\noindent
with {\small $w_p=1$},
{\small $w_q=7.608\times10^{-2}$},
{\small $w_{\mathrm{fk}}=7.109\times10^{-2}$},
{\small $w_{\ell}=3.733\times10^{-1}$},
{\small $w_{\mathrm{eq}}=5.998\times10^{-1}$}, and
{\small $w_{\mathrm{con}}=2.028\times10^{-3}$}.
Here, {\small $\mathcal{L}_{\mathrm{fk}}$} enforces consistency between the predicted marker positions and the GVS forward kinematics, {\small $\mathcal{L}_{\ell}$} enforces tendon length consistency, {\small $\mathcal{L}_{\mathrm{phys}}$} represents the projected static equilibrium loss, and {\small $\mathcal{L}_{\mathrm{con}}$} represents the closed chain geometric constraint loss. On the simulation test set, the pretrained network achieved a marker RMSE of {\small $0.164~\mathrm{mm}$} and a position coefficient of determination of {\small $R^2=0.9987$}. The resulting network was subsequently used to initialize the experimental fine tuning.

%======================================================================
\subsection{Experimental Fine Tuning}
\label{subsec:experimental_finetuning}
%======================================================================

The {\small $817$} experimental configurations were divided into
{\small $571$} training, {\small $122$} validation, and
{\small $124$} test samples. Repeated measurements corresponding to
the same actuation input were assigned to the same partition to prevent
information leakage. The test set was excluded from fine tuning and
model selection.

The tendon displacement inputs and marker displacement targets were
standardized using the normalization statistics obtained from the
simulation training set, ensuring the same scaling used during
simulation pretraining.

Because the GVS generalized coordinates cannot be measured directly
in the experiment, no experimental labels are available for
{\small $\mathbf{q}$}. A frozen copy of the simulation pretrained PINN
was therefore used as a reference model to provide auxiliary
generalized coordinate predictions. For each standardized experimental
input, the reference prediction is
{\small
$
\widetilde{\mathbf{q}}_{\mathrm{ref}}^{(j)}
=
\mathcal{N}_{\boldsymbol{\theta}_{\mathrm{sim}}}^{q}
\left(
\widetilde{\Delta\boldsymbol{\ell}}^{(j)}
\right),
$
}
\noindent
where {\small $\boldsymbol{\theta}_{\mathrm{sim}}$} denotes the
parameters of the simulation pretrained PINN. The generalized
coordinate consistency loss is defined as
{\small
$
\mathcal{L}_{q,\mathrm{ref}}
=
\frac{1}{16N_b}
\sum_{j=1}^{N_b}
\left\|
\widehat{\widetilde{\mathbf{q}}}^{(j)}
-
\widetilde{\mathbf{q}}_{\mathrm{ref}}^{(j)}
\right\|_2^2,
$
}
\noindent
where {\small $N_b$} denotes the mini batch size. The experimental
marker displacement loss is evaluated in standardized space as
{\small
$
\mathcal{L}_{p,\mathrm{exp}}
=
\frac{1}{24N_b}
\sum_{j=1}^{N_b}
\left\|
\Delta\widehat{\widetilde{\mathbf{p}}}^{(j)}
-
\Delta\widetilde{\mathbf{p}}_{\mathrm{exp}}^{(j)}
\right\|_2^2.
$
}
During fine tuning, the calibrated loss weights obtained from
simulation pretraining were retained. The contribution of the physics
terms was reduced using
{\small $\alpha_{\mathrm{phys}}=0.05$} to allow adaptation to the
experimental measurements while retaining the physical constraints.
For mini batches in which the physics terms were evaluated, the
training objective was

{\small
\begin{equation}
\begin{aligned}
\mathcal{L}_{\mathrm{FT}}
={}&
w_p\mathcal{L}_{p,\mathrm{exp}}
+
w_q\mathcal{L}_{q,\mathrm{ref}} \\
&+
\alpha_{\mathrm{phys}}
\left(
w_{\mathrm{fk}}\mathcal{L}_{\mathrm{fk}}
+
w_{\ell}\mathcal{L}_{\ell}
+
w_{\mathrm{eq}}\mathcal{L}_{\mathrm{phys}}
+
w_{\mathrm{con}}\mathcal{L}_{\mathrm{con}}
\right).
\end{aligned}
\label{eq:experimental_finetuning_loss}
\end{equation}
}

\noindent
The weights {\small $w_p$}, {\small $w_q$},
{\small $w_{\mathrm{fk}}$}, {\small $w_{\ell}$},
{\small $w_{\mathrm{eq}}$}, and {\small $w_{\mathrm{con}}$}
are the fixed values obtained during simulation pretraining. The
reduced physics contribution allows the network to adapt to the
measured marker configurations while preserving consistency with the
GVS forward kinematics, tendon displacements, static equilibrium, and
closed chain constraints.

Fine tuning was performed using AdamW with a mini batch size of
{\small $64$}, weight decay of {\small $10^{-6}$}, and gradient norm
clipping at {\small $2$}. The physics terms were evaluated for two
samples in one out of every three mini batches. For the remaining
mini batches, training used only the experimental marker displacement
loss and generalized coordinate consistency loss.

The model was first fine tuned for {\small $100$} epochs using a
learning rate of {\small $5\times10^{-5}$}. Training was then
continued for up to {\small $50$} additional epochs with a reduced
learning rate of {\small $10^{-5}$}. Model selection was based
exclusively on the validation marker RMSE. The selected model was
obtained at epoch {\small $149$}, with a validation marker RMSE of
{\small $0.4921~\mathrm{mm}$}.

\vspace{-0.5 em}
%======================================================================
\subsection{Experimental Evaluation and Results}
\label{subsec:experimental_results}
%======================================================================

The experimental performance was evaluated on the
{\small $124$} test configurations that were excluded from fine tuning
and model selection. The simulation pretrained PINN and the final fine
tuned model were evaluated on the same test set.

For test sample {\small $j$} and marker {\small $m$}, the Euclidean
marker error is defined as
{\small
$
e_{j,m}
=
\left\|
\widehat{\mathbf{p}}_{j,m}
-
\mathbf{p}_{j,m}
\right\|_2,
$
}
\noindent
where {\small $\widehat{\mathbf{p}}_{j,m}$} and
{\small $\mathbf{p}_{j,m}$} denote the predicted and measured
three dimensional marker positions, respectively. The overall marker
RMSE is calculated as
{\small
$
\mathrm{RMSE}_{m}
=
\sqrt{
\frac{1}{8N_t}
\sum_{j=1}^{N_t}
\sum_{m=1}^{8}
e_{j,m}^{2}
},
$
}
\noindent
where {\small $N_t$} denotes the number of test configurations.
The mean and maximum marker errors, coefficient of determination
{\small $R^2$}, and percentage of marker predictions within
{\small $1~\mathrm{mm}$} of the Vicon measurements are also reported.

Table~\ref{tab:experimental_results} summarizes the prediction
performance on the held out experimental test set before and after fine
tuning. The marker RMSE decreases from
{\small $2.657~\mathrm{mm}$} to {\small $0.497~\mathrm{mm}$},
corresponding to an {\small $81.3\%$} reduction, while the mean marker
error decreases from {\small $1.981~\mathrm{mm}$} to
{\small $0.382~\mathrm{mm}$}. The maximum marker error is reduced from
{\small $9.244~\mathrm{mm}$} to {\small $2.282~\mathrm{mm}$}.
Moreover, {\small $R^2$} increases from {\small $-0.788$} to
{\small $0.937$}, and the percentage of marker predictions within
{\small $1~\mathrm{mm}$} of the measured positions increases from
{\small $37.30\%$} to {\small $94.96\%$}.

\begin{table}[!t]
\caption{Experimental Performance Before and After Fine Tuning}
\label{tab:experimental_results}
\centering
\footnotesize
\setlength{\tabcolsep}{2.5pt}
\renewcommand{\arraystretch}{1.15}

\begin{tabular}{lccccc}
\toprule
\textbf{Model} &
\textbf{RMSE} &
\textbf{Mean Error} &
\textbf{Max. Error} &
\textbf{$R^2$} &
\textbf{$\leq1$ mm} \\
&
\textbf{(mm)} &
\textbf{(mm)} &
\textbf{(mm)} &
&
\textbf{(\%)} \\
\midrule

Before fine tuning
& 2.657
& 1.981
& 9.244
& -0.788
& 37.30 \\

After fine tuning
& \textbf{0.497}
& \textbf{0.382}
& \textbf{2.282}
& \textbf{0.937}
& \textbf{94.96} \\

\bottomrule
\end{tabular}
\end{table}

\begin{figure}[!t]
    \centering
    \includegraphics[
        width=0.65\columnwidth
    ]{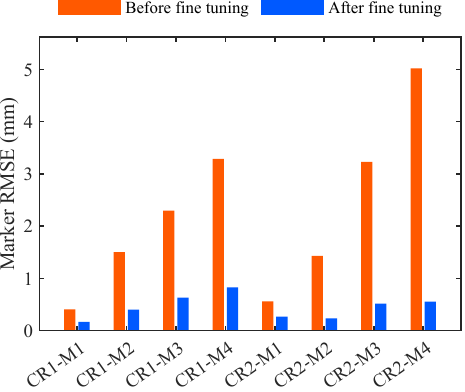}
    \caption{Marker specific RMSE before and after fine tuning at the
    four measured locations on each continuum robot.}
    \label{fig:experimental_before_after}
\end{figure}

Figure~\ref{fig:experimental_before_after} shows the marker specific
RMSE along the two CRs. Fine tuning reduces the prediction
error at all eight measured locations. After fine tuning, the largest
marker RMSE is {\small $0.827~\mathrm{mm}$}, observed at CR1\_M4,
with all remaining marker errors below this value. This indicates that
the improvement is distributed across the measured CCR configuration
rather than being limited to particular marker locations.

\begin{figure}[!t]
    \centering
    \includegraphics[
        width=0.95\columnwidth
    ]{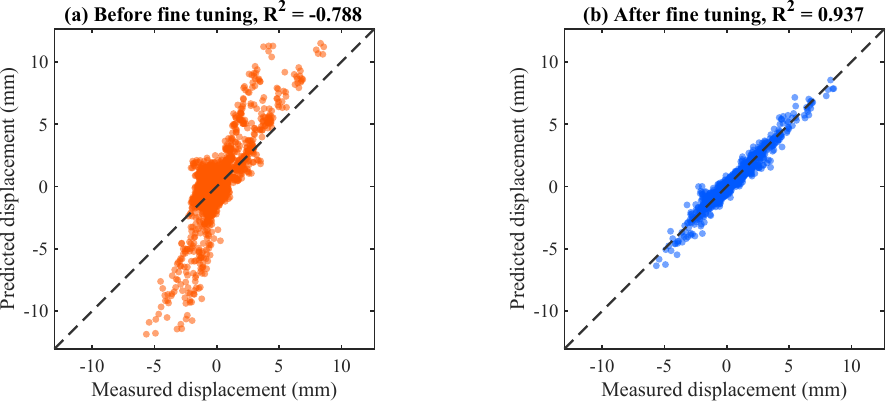}
    \caption{Measured versus predicted marker displacements on the
    held out experimental test set (a) before and (b) after fine
    tuning. The dashed line represents ideal agreement between the
    measured and predicted values.}
    \label{fig:measured_predicted_exp}
\end{figure}

Figure~\ref{fig:measured_predicted_exp} further compares the measured
and predicted marker displacements. Before fine tuning, the predictions
show substantial deviations from the ideal agreement line. After fine
tuning, the predicted displacements closely follow the measured values
over the experimental range, consistent with the increase in
{\small $R^2$} from {\small $-0.788$} to {\small $0.937$}. These
results demonstrate that experimental fine tuning substantially reduces
the discrepancy between the simulation pretrained model and the
physical CCR while retaining the physics informed formulation.

%======================================================================
\section{Conclusion}
\label{sec:conclusion}
%======================================================================

This paper presented a constraint-aware PINN for static shape estimation
of closed-chain CCRs. The proposed formulation incorporates the GVS
static model through a projected equilibrium residual that eliminates
the unknown closed-chain reaction forces, together with a geometric
residual that enforces loop closure. Simulation results showed that the
physics informed formulation improves robustness to limited and noisy
training data while producing more mechanically consistent predictions
than a purely data driven ANN. The learned static mapping also provides
a substantial computational advantage over iterative nonlinear
equilibrium solution while maintaining high prediction accuracy.
Experimental validation further showed that fine tuning the simulation
pretrained network using measured marker data effectively reduces the
simulation to real discrepancy and improves agreement with the physical
CCR.
Future work will extend the proposed framework to dynamic modeling and
time varying tendon inputs, where inertial and damping effects must be
included in the physics informed formulation. Further investigation will
also consider online adaptation to model uncertainty and changing robot
properties, integration of the learned model into closed loop control
and optimization, and extension to CCR systems with different
geometries, actuation arrangements, and manipulated objects.

\bibliographystyle{IEEEtran}
\bibliography{references_cleaned}

\end{document}